\documentclass[10pt,twocolumn,letterpaper]{article}
\usepackage{colortbl}
\usepackage{cvpr}
\usepackage{times}
\usepackage{epsfig}
\usepackage{graphicx}
\usepackage{amsmath}
\usepackage{amssymb}
\usepackage{xspace}
\usepackage{stackrel}
\usepackage[pagebackref=true,breaklinks=true,letterpaper=true,colorlinks,bookmarks=false]{hyperref}

\newcommand{\G}{\mathbf{G}}

\newcommand{\W}{\mathbf{W}}
\newcommand{\X}{\mathbf{X}}

\newcommand{\f}{\mathbf{f}}

\newcommand{\w}{\mathbf{w}}
\newcommand{\x}{\mathbf{x}}

\newcommand{\cX}{\mathcal{X}}

\newcommand{\real}{\mathbb{R}}
\newcommand{\bina}{\{0, 1\}}

\newcommand{\udots}{\mathinner{\mskip1mu\raise1pt\vbox{\kern7pt\hbox{.}}\mskip2mu\raise4pt\hbox{.}\mskip2mu\raise7pt\hbox{.}\mskip1mu}}

\newenvironment{aligns}{\par\nobreak\small\noindent\align}{\endalign}

\cvprfinalcopy
\def\cvprPaperID{1036} 

\ifcvprfinal\fi
\begin{document}

\title{Fine-grained Image Classification by Exploring Bipartite-Graph
  Labels}

\author{Feng Zhou\\
NEC Labs\\
{\tt\small www.f-zhou.com}
\and
Yuanqing Lin\\
NEC Labs\\
{\tt\small ylin@nec-labs.com}
}

\maketitle

\begin{abstract}
  Given a food image, can a fine-grained object recognition engine
  tell ``which restaurant which dish'' the food belongs to? Such
  ultra-fine grained image recognition is the key for many
  applications like search by images, but it is very challenging
  because it needs to discern subtle difference between classes while
  dealing with the scarcity of training data. Fortunately, the
  ultra-fine granularity naturally brings rich relationships among
  object classes. This paper proposes a novel approach to exploit the
  rich relationships through bipartite-graph labels (BGL). We show how
  to model BGL in an overall convolutional neural networks and the
  resulting system can be optimized through back-propagation. We also
  show that it is computationally efficient in inference thanks to the
  bipartite structure. To facilitate the study, we construct a new
  food benchmark dataset, which consists of 37,885 food images
  collected from 6 restaurants and totally 975 menus. Experimental
  results on this new food and three other datasets demonstrates BGL
  advances previous works in fine-grained object recognition. An
  online demo is available at {\tt\url{http://www.f-zhou.com/fg_demo/}}.
\end{abstract}

\vspace{-.2in}

\section{Introduction}

Fine-grained image classification concerns the task of distinguishing
sub-ordinate categories of some base classes such as
dogs~\cite{KhoslaJYF11,ParkhiVZJ12},
birds~\cite{BergLLAJB14,BransonHWPB14},
flowers~\cite{AngelovaZ13,NilsbackZ08},
plants~\cite{KumarBBJKLS12,SfarBG13},
cars~\cite{KrauseSDF013,LinMHD14,StarkKPMLSK12},
food~\cite{BeijbomJMSK15,BossardGG14,MyersJRKGSGPHM15,YangCPS10},
clothes~\cite{DiWBPS13}, fonts~\cite{ChenYJBSAH14} and
furniture~\cite{BellB15}. It differs from the base-class
classification~\cite{EveringhamEGWWZ15} in that the differences among
object classes are more subtle, and thus it is more difficult to
distinguish them. Yet fine-grained object classification is extremely
useful because it is the key to a variety of challenging applications
even difficult for human annotators.

\begin{figure}
  \centering
  \includegraphics[width=.5\textwidth]{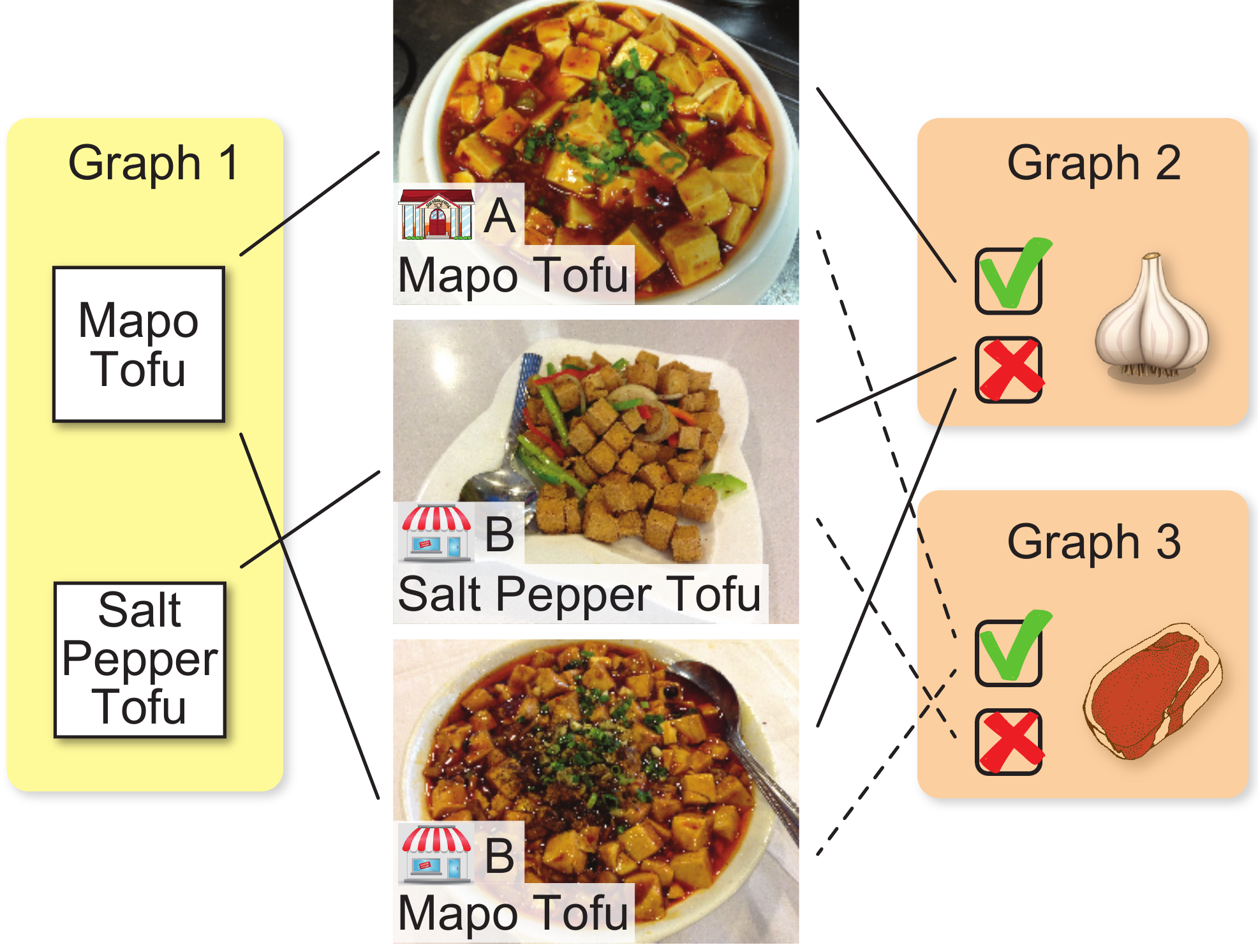}
  \caption{Illustration of three ultra-fine grained classes (middle),
    Mapo Tofu of Restaurant A, Salt Pepper Tofu of Restaurant B, and
    Mapo Tofu of Restaurant B. Their relationships can be modeled
    through three bipartite graphs, fine-grained classes vs. general
    food dishes (left) and fine-grained classes vs. two ingredients
    (right). This paper shows how to incorporate the rich
    bipartite-graph labels (BGL) into convolutional neural network
    training to improve recognition accuracy.}
  \label{fig:over}
\end{figure}

While general image classification has achieved impressive success
within the last few
years~\cite{KrizhevskySH12,SzegedyLJSRAEVR14,SimonyanZ14a,HeZR015,WuYSDS15,IoffeS15},
it is still very challenging to recognize object classes with
ultra-fine granularity. For instance, how to recognize each of the
three food images shown in Fig.~\ref{fig:over} into \emph{which
  restaurant which dish}? The challenge arises in two major
aspects. First, different classes may be visually similar, \eg, the
Mapo Tofu dish from Restaurant A (the $1^{st}$ image in
Fig.~\ref{fig:over}) looks very similar to the one from Restaurant B
(the $3^{rd}$ image in Fig.~\ref{fig:over}). Second, each class may
not have enough training images because of the ultra-fine
granularity. In such setting, how to share information between similar
classes while maintaining strong discriminativeness becomes more
critical.


To that end, we propose a novel approach using bipartite-graph labels
(BGL) that models the rich relationships among the ultra-fine grained
classes. In the example of Fig.~\ref{fig:over}, the $1^{st}$ and
$3^{rd}$ images are both Mapo Tofu dishes; and they share some
ingredients with the $2^{nd}$ one. Such class relationships can be
modeled in three bipartite graphs. This paper shows how to incorporate
the class bipartite graphs into CNN and learn the optimal classifiers
through overall back-propagation.

Using BGL has several advantages: (1) BGL imposes additional
constraints to regularize CNN training, thereby largely reducing the
possibility of being overfitting when only a small amount of training
data is available. (2) Knowing classes that belong to the same coarse
category or share some common attributes can allow us to borrow some
knowledge from relevant classes. (3) The supervised feature learning
through a global back-propagation allows learning discriminative
features for capturing subtle differences between similar classes. (4)
By constraining the structure to bipartite graphs, BGL prevents the
exponential explosion from enumerating all possible states in
inference.

This work is in parallel to the existing big body of fine-grained
image classification research, which has focused on devising more
discriminative feature by aligning object
poses~\cite{FarrellOZMDD11,GavvesFSST13,ZhangDGD14} and filtering out
background through object segmentation~\cite{ParkhiVJZ11}. The
techniques developed in this work can be combined with the ones in
those existing research works to achieve better fine-grained image
recognition performance.

To facilitate the study, we built an ultra-fine grained image
recognition benchmark, which consists of 37,885 food training images
collected directly from 6 restaurants with totally 975 menus. Our
results show that the proposed BGL approach produces significantly
better recognition accuracy compared to the powerful
GoogLeNet~\cite{SzegedyLJSRAEVR14}. We also test the BGL approach on
some existing fine-grained benchmark datasets. We observe the benefit
of BGL modeling as well although the improvement is less significant
because of the less rich class relationships.



\section{Previous work}

This section reviews the related work on fine-grained image
classification, and structural label learning.

\subsection{Fine-grained image classification}




Fine-grained image classification needs to discern subtle differences
among similar classes. The majority of existing approaches have thus
been focusing on localizing and describing discriminative object parts
in fine-grained domains. Various pose-normalization pooling strategies
combined with 2D or 3D geometry have been proposed for recognizing
birds~\cite{BransonHPB14,FarrellOZMDD11,GavvesFSST13,ZhangDGD14,ZhangFID13,ZhangPRDB14},
dogs~\cite{LiuKJB12} and cars~\cite{KrauseGDLF14,KrauseSDF013}. The
main drawback of these approaches is that part annotations are
significantly more challenging to collect than image labels. Instead,
a variety of methods have been developed towards the goal of finding
object parts in an unsupervised or semi-supervised fashion. Krause
\etal~\cite{KrauseGDLF15} combined alignment with co-segmentation to
generating parts without annotations. Lin \etal.~\cite{LinCM15}
proposed an architecture that uses two separate CNN feature extractors
to model the appearance due to where the parts are and what the parts
look like. Jaderberg \etal.~\cite{JaderbergSZK15} introduced spatial
transformer, a new differentiable module that can be inserted into
existing convolutional architectures to spatially transform feature
maps without any extra training supervision. In parallel to above
efforts, our approach focuses on exploiting rich class relationships
and is applicable to generic fine-grained objects even they do not own
learnable structures (\eg, food dishes).

To provide good features for recognition, another prominent direction
is to adopt detection and segmentation methods as an initial step and
to filter out noise and clutter in background. For instance, Parkhi
\etal~\cite{ParkhiVJZ11,ParkhiVZJ12} proposed to detect some specific
part (\eg, cat's head) and then performed a full-object segmentation
through propagation. In another similar work, Angelova and
Zhu~\cite{AngelovaZ13} further re-normalized objects after
segmentation to improves the recognition performance. However, better
feature through segmentation always comes with computational cost as
segmentation is often computationally expensive.


Recently, many other advances lead to improved results. For instance,
Wang \etal.~\cite{WangSLRWPCW14} and Qian \etal.~\cite{QianJZL15}
showed a more discriminative similarity can be learned through deep
ranking and metric learning respectively. Xie \etal.~\cite{XieYWL15}
proposed a novel data augmentation approach to better regularize the
fine-grained problems. Deng \etal.~\cite{DengK013}, Branson
\etal.~\cite{BransonHWPB14} and Wilber \etal~\cite{WilberKKB15}
developed hybrid systems to introduce human in the loop for localizing
discriminative regions for computing features. Focusing on different
goal on exploring label structures, our method can be potentially
combined with the above methods to further improve the performance.

\subsection{Structural label learning}

While most existing works focus on single-label classification
problem, it is more natural to describe real world images with
multiple labels like tags or attributes. According to the assumptions
on label structures, previous work on structural label learning can be
roughly categorized as learning binary, relative or hierarchical
attributes.

Much of prior work focuses on learning binary attributes that indicate
the presence of a certain property in an image or not. For instance,
previous works have shown the benefit of learning binary attributes
for face verification~\cite{KumarBBN11}, texture
recognition~\cite{FerrariZ07}, clothing searching~\cite{DiWBPS13}, and
zero-shot learning~\cite{LampertNH09}. However, binary attributes are
restrictive when the description of certain object property is
continuous or ambiguous.

To address the limitation of binary attributes, comparing attributes
has gained attention in the last years. The relative-attribute
approach~\cite{ParikhG11} learns a global linear ranking function for
each attribute, offering a semantically richer way to describe and
compare objects. While a promising direction, a global ranking of
attributes tends to fail when facing fine-grained visual
comparisons. Yu and Grauman~\cite{YuG14} fixed this issue by learning
local functions that tailor the comparisons to neighborhood statistics
of the data. Recently, Yu and Grauman~\cite{YuG15} developed a
Bayesian strategy to infer when images are indistinguishable for a
given attribute.

Our method falls into the third category where the relation between
the fine-grained labels and attributes is modeled in a hierarchical
manner. In the past few years, extensive research has been devoted to
learning a hierarchical structure over classes (see~\cite{TouschHA12}
for a survey). Previous works have shown the benefits of leveraging
the semantic class hierarchy using either
unsupervised~\cite{BartPPW08,TodorovicA08} or
supervised~\cite{DengBLF10,FergusBWT10,NamMKF15} methods. Our work
differs from previous works in the CNN-based framework and the setting
focusing on multi-labeled object. The most similar works to ours are
\cite{SrivastavaS13} and \cite{DengDJFMBLNA14}, which show the
advantages of exploring the tree-like hierarchy in small-scale (\eg,
CIFAR-100) and graph-like structure in large-scale (\eg, ImageNet)
categories respectively. Compared to \cite{SrivastavaS13}, our method
is able to handle more general structure (\eg, attributes) among
fine-grained labels. Unlike \cite{DengDJFMBLNA14} relying on
approximated inference, our method allows for efficient exact
inference by modeling the label dependence as a star-like combination
of bipartite graphs. In addition, we explore the hierarchical
regularization on the last fully connected layer, which could further
reduce the possibility of being overfitting.

Our work is also related to previous methods in multi-task learning
(MTL)~\cite{Caruana97}. To better learn multiple correlated subtasks,
previous work have explored various ideas including, sharing hidden
nodes in neural networks~\cite{Caruana97}, placing a common prior in
hierarchical Bayesian models~\cite{XueLCK07} and structured
regularization in kernel methods~\cite{EvgeniouP04}, among others. Our
method differs from the MTL methods mainly in the setting of
multi-label setting for single objects.

\section{CNN with Bipartite-Graph Labels}

This section describes the proposed BGL method in a common multi-class
convolutional neural network (CNN) framework, which is compatible to
most popular architectures like, AlexNet~\cite{KrizhevskySH12},
GoogLeNet~\cite{SzegedyLJSRAEVR14} and VGGNet~\cite{SimonyanZ14a}. Our
approach modifies their softmax layer as well as the last fully
connected layer to explicitly model BGLs. The resulted CNN is
optimized by a global back-propagation.

\subsection{Background}

Suppose that we are given (see notation\footnote{Bold capital letters
  denote a matrix $\X$, bold lower-case letters a column vector
  $\x$. All non-bold letters represent scalars. $\x_i$ represents the
  $i^{th}$ column of the matrix $\X$. $x_{ij}$ denotes the scalar in
  the $i^{th}$ row and $j^{th}$ column of the matrix $\X$.
  $1_{[i = j]}$ is an indicator function and its value equals to $1$
  if $i = j$ and $0$ otherwise.}) a set of $n$ images
$\cX = \{ (\x, y), \cdots \}$ for training, where each image $\x$ is
annotated with one of $k$ fine-grained labels, $y = 1, \cdots, k$. Let
$\x \in \real^d$ denote the input feature of the last fully-connected
layer, which generates $k$ scores $\f \in \real^{k}$ through a linear
function $\f = \W^T \x$ defined by the parameters
$\W \in \real^{d \times k}$. In a nutshell, the last layer of CNN is
to minimize the negative log-likelihood over the training data, \ie,
\begin{aligns}
  \min_{\W} \ \sum_{(\x, y) \in \cX} -\log P(y \big| \x,
  \W), \label{eq:obj_ori}
\end{aligns}
where the softmax score,
\begin{aligns}
  P(i \big| \x, \W) = \frac{e^{f_i}}{\sum_{j=1}^{k}
    e^{f_j}} \doteq p_i, \label{eq:soft}
\end{aligns}
encodes the posterior probability of image $\x$ being classified as
the $i^{th}$ fine-grained class.

\subsection{Objective function with BGL modeling}
\label{sec:obj}

Despite the great improvements achieved on base-class recognition in
the last few years, recognizing object classes in ultra-fine
granularity like the example shown in Fig.~\ref{fig:over} is still
challenging mainly for two reasons. First, unlike general recognition
task, the amount of labels with ultra-fine granularity is often
limited. The training of a high-capacity CNN model is thus more prone
to overfitting. Second, it is difficult to learn discriminative
features to differentiate between similar fine-grained classes in the
presence of large within-class difference.

To address these difficulties, we propose bipartite-graph labels (BGL)
to jointly model the fine-grained classes with pre-defined
\textit{coarse classes}. Generally speaking, the choices of coarse
classes can be any grouping of fine-grained classes. Typical examples
include bigger classes, attributes or tags. For instance,
Fig.~\ref{fig:over} shows three types of coarse classes defined on the
fine-grained Tofu dishes (middle). In the first case (Graph 1 in
Fig.~\ref{fig:over}), the coarse classes are two general Tofu food
classes by neglecting the restaurant tags. In the second and third
cases (Graph 2 and 3 in Fig.~\ref{fig:over}), the coarse classes are
binary attributes according to the presence of some ingredient in the
dishes. Compared to the original softmax loss (Eq.~\ref{eq:soft})
defined only on fine-grained labels, the introduction of coarse
classes in BGL has three benefits: (1) New coarse classes bring in
additional supervised information so that the fine-grained classes
connected to the same coarse class can share training data with each
other. (2) All fine-grained classes are implicitly ranked according to
the connection with coarse classes. For instance, Toyota Camry 2014
and Toyota Camry 2015 are much closer to each other compared to Honda
Accord 2015. This hierarchical ranking guides CNN training to learn
more discriminative features to capture subtle difference between
fine-grained classes. (3) Compared to image-level labels (\eg, image
attribute, bounding box, segmentation mask) that are expensive to
obtain, it is relatively cheaper and more natural to define coarse
classes over fine-grained labels. This fact endows BGL the board
applicability in real scenario.

Given $m$ types of coarse classes, where each type $j$ contains $k_j$
coarse classes, BGL models their relations with the $k$ fine-grained
classes as $m$ bipartite graphs grouped in a star-like structure. Take
Fig.~\ref{fig:over} for instance, where the three types of coarse
classes form three separated bipartite graphs with the fine-grained
Tofu dishes, and there is no direct connection among the three types
of coarse classes. For each graph of coarse type $j$, we encode its
bipartite structure in a binary association matrix $\G_j \in \bina^{k
  \times k_j}$, whose element $g^j_{ic_j} = 1$ if the $i^{th}$
fine-grained label is connected with coarse label $c_j$.  As it will
become clear later, this star-like composition of bipartite graphs
enables BGL to perform exact inference as opposed to the use of other
general label graphs (\eg, \cite{DengDJFMBLNA14}).

To generate the scores $\f_j = \W_j^T \x \in \real^{k_j}$ for coarse
classes of type $j$, we augment the last fully-connected layer with
$m$ additional variables, $\{\W_j\}_j$, where $\W_j \in \real^{d
  \times k_j}$. Given an input image $\x$ of $i^{th}$ fine-gained
class, BGL models its joint probability with any $m$ coarse labels
$\{c_j\}_j$ as,
\begin{aligns}
  P(i, \{ c_j \}_j \big| \x, \W, \{ \W_j \}_j) &=
  \frac{1}{z} e^{f_i} \prod_{j=1}^m g^j_{ic_j} e^{f^j_{c_j}}, \nonumber
\end{aligns}
where $z$ is the partition function computed as:
\begin{aligns}
  z & = \sum_{i=1}^k e^{f_i} \prod_{j=1}^m \sum_{c_j=1}^{k_j} g^j_{i c_j} e^{f^j_{c_j}}. \nonumber
\end{aligns}
At first glance, computing $z$ is infeasible in practice. Because of
the bipartite structure of the label graph, however, we could denote
the non-zero element in $i^{th}$ row of $\G_j$ as $\phi_{i}^j = c_j$
where $g^j_{i c_j} = 1$. With this auxiliary function, the computation
of $z$ can be simplified as
\begin{aligns}
  z & = \sum_{i=1}^k e^{f_i} \prod_{j=1}^m e^{f^j_{\phi_{i}^j}}. \label{eq:z2}
\end{aligns}
Compared to general CRF-based methods (\eg, \cite{DengDJFMBLNA14})
with exponential number of possible states, the complexity $O(k m)$ of
computing $z$ in BGL through Eq.~\ref{eq:z2} scales linearly with
respect to the number of fine-grained classes ($k$) as well as the
type of coarse labels ($m$). Given $z$, the marginal posterior
probability over fine-grained and coarse labels can be computed as:
\begin{aligns}
  P(i \big| \x, \W, \{ \W_j \}_j) &= \frac{1}{z} e^{f_i} \prod_{j=1}^m e^{f^j_{\phi^j_i}} \doteq p_i ,
  \nonumber \\
  P(c_j \big| \x, \W, \{ \W_l \}_l) &= \frac{1}{z} \sum_{i=1}^k g^j_{i c_j} e^{f_i} \prod_{l=1}^m e^{f^l_{\phi^l_i}}
  \doteq p_{c_j}^j. \nonumber
\end{aligns}



As discussed before, one of the difficulties in training CNN is the
possibility of being overfitting. One common solution is to add a
$l_2$ weight decay term, which is equivalent to sample the columns of
$\W$ from a Gaussian prior. Given the connection among fine-grained
and coarse classes, BGL provides another natural hierarchical prior
for sampling the weights:
\begin{aligns}
  P(\W, \{ \W_j \}_j) &= \prod_{i=1}^k \prod_{j=1}^m \prod_{c_j=1}^{k_j} e^{- \frac{\lambda}{2} g^j_{i c_j} \| \w_i -
    \w^j_{c_j} \|^2} \doteq p_w. \nonumber
\end{aligns}
This prior expects $\w_i$ and $\w_{c_j}^j$ have small distance if
$i^{th}$ fine-grained label is connected to coarse class $c_j$ of type
$j$. Notice that this idea is a generalized version of the one
proposed in \cite{SrivastavaS13}. However, \cite{SrivastavaS13} only
discussed a special type of coarse label (big class), while BGL can
handle much more general coarse labels such as multiple attributes.


In summary, given the training data $\cX$ and the graph label defined
by $\{\G_j\}_j$, the last layer of CNN with BGL aims to minimize the
joint negative log-likelihood with proper regularization over the
weights:
\begin{aligns} \label{eq:obj_bgl}
  \min_{\W, \{\W_j\}_j} \ & \sum_{(\x, y) \in \cX} \Big( - \log
  p_{y} - \sum_{j=1}^m \log p_{\phi^j_{y}}^j \Big) - \log
  p_w.
\end{aligns}

\subsection{Optimization}

We optimized BGL using the standard back-propagation with mini-batch
stochastic gradient descent. The gradients for each parameter can be
computed\footnote{See supplementary materials for detailed
  derivation.} all in closed-form:
\begin{aligns}
  \frac{\partial \log p_y}{\partial f_i} &= 1_{[i =
    y]} - p_{i}, \ \frac{\partial \log p_y}{\partial f^j_{c_j}} = 1_{[ g_{yc_j}^j = 1]} - p_{c_j}^j, \nonumber \\
  \frac{\partial \log p^j_{\phi^j_y }}{\partial f_i} &= \frac{p_i}{
    p^j_{\phi^j_y} } 1_{[g^j_{i \phi_y^j} = 1]} - p_{i}, \
  \frac{\partial \log p^j_{\phi^j_y }}{\partial f_{c_j}^j} = 1_{[c_j =
    \phi^j_y]} - p_{c_j}^j, \nonumber \\
  \frac{\partial \log p^j_{\phi^j_y }}{\partial f_{c_l}^l} &=
  \sum_{i=1}^k g^j_{i \phi^j_y} g^l_{i l} \frac{p_i}{p^j_{\phi^j_y}} -
  p^l_{c_l}, l \neq j, \label{eq:pij} \\
  \frac{\partial \log p_w}{\partial \w_i} &= -\lambda \sum_{j=1}^m
  \sum_{c_j=1}^{k_j} g^j_{i c_j}
  (\w_i - \w_{c_j}^{j}), \label{eq:opt_w1} \\
  \frac{\partial \log p_w}{ \partial \w_{c_j}^j} &= -\lambda
  \sum_{i=1}^k g^j_{i c_j} (\w_{c_j}^j - \w_i). \label{eq:opt_w2}
\end{aligns}

\begin{figure}
  \centering
  \includegraphics[width=.5\textwidth]{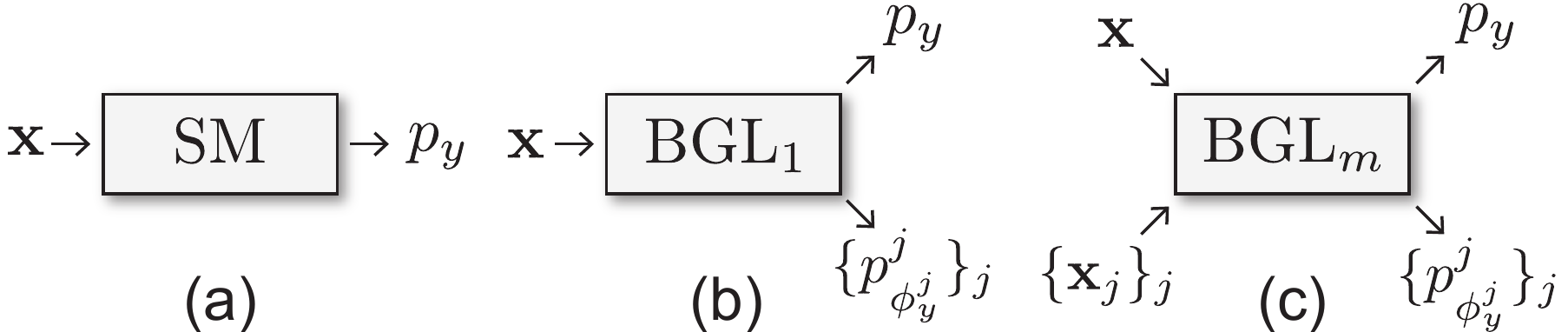}
  \caption{Comparison of output layers. (a) Softmax (SM). (b) BGL with
    one feature (BGL$_1$). (c) BGL with multiple features (BGL$_m$).}
  \label{fig:conf}
\end{figure}

Here we briefly discuss some implementation issues. (1) Computing $p_i
/ p^j_{\phi^j_y}$ by independently calculating $p_i$ and
$p^j_{\phi^j_y}$ is not numerically stable because $p^j_{\phi^j_y}$
could be very small. It is better to jointly normalize the two terms
first. (2) Directly computing Eq.~\ref{eq:pij} has a quadratic
complexity with respect to the number of coarse classes. But it can be
reduced to linear because most computations are redundant. See
supplementary materials for more details. (3) So far
(Fig.~\ref{fig:conf}b) we assume the same feature $\x$ is used for
computing both the fine-grained $\f = \W^T \x$ and coarse scores $\f_j
= \W_j^T \x$. In fact, BGL can naturally combine multiple CNNs as
shown in Fig.~\ref{fig:conf}c. This allows the model to learn
different low-level features $\x_j$ for coarse labels $\f_j = \W_j^T
\x_j$.

\section{Experiments}

\begin{figure*}
  \centering
  \includegraphics[width=\textwidth]{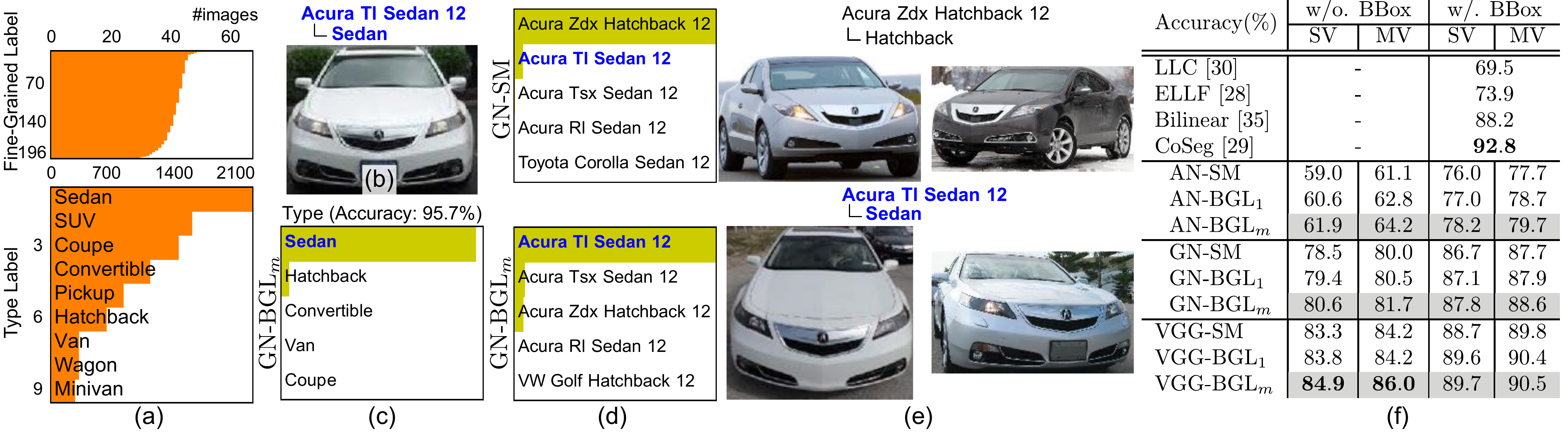}
  \caption{Comparison on the Stanford car dataset. (a) The number of
    training images for each fine-grained (top) and coarse (bottom)
    category. (b) An example of testing images. (c) Type predicted by
    GN-BGL$_m$. (d) Top-5 predictions generated by GN-SM approach (top)
    and the proposed BGL approach GN-BGL respectively. (e) Similar
    training exemplars according to the input feature $\x$. (f)
    Accuracy.}
  \label{fig:exp_carsf}
\end{figure*}

This section evaluates BGL's performance for fine-grained object
recognition on four benchmark datasets. The first two datasets,
Stanford cars~\cite{KrauseSDF013}, CUB-200-2011~\cite{WahCBWPB11},
contains fine-grained categories sharing one level of coarse
attributes. The last two datasets, Car-333~\cite{XieYWL15} and a new
Food-975 dataset, consist of ultra-fine grained classes and richer
class relationships. We wish the proposed BGL approach would be able
to improve classification accuracy even with simple BGLs and
significantly improve performance when richer class relationships are
present.

BGL was implemented on the off-the-shelf Caffe~\cite{JiaSDKLGGD14}
platform. We test on three popular CNN frameworks, AlexNet
(\textbf{AN})~\cite{KrizhevskySH12}, GoogLeNet
(\textbf{GN})~\cite{SzegedyLJSRAEVR14}, and VGGNet
(\textbf{VGG})~\cite{SimonyanZ14a}. For each of them, we compared
three settings: \textbf{SM} with the traditional softmax loss;
\textbf{BGL$_1$} by modifying the softmax layer and the last fully
connected layer as the proposed BGL approach; and \textbf{BGL$_m$} by
combining multiple networks\footnote{Limited by the GPU memory, we
  always combined two networks in BGL$_m$, one for fine-grained
  classes and the other for all coarse labels.} that generate
different features for fine-grained and coarse classes.

Our models were trained for $100$ epochs on a single NVIDIA K40
GPU. We adopted the default hyper-parameters as used by Caffe. In all
experiments, we fine-tuned from pre-trained ImageNet model as in
\cite{KarayevTHADHW14} because it always achieved better
result. During training, we down-sampled the images to a fixed
$256$-by-$256$ resolution, from which we randomly cropped
$224$-by-$224$ patches for training. We also did their horizontal
reflection for further data augmentation. During testing, we evaluated
the top-1 accuracy using two cropping strategies: (1) single-view
(\textbf{SV}) by cropping the center $224$-by-$224$ patch of the
testing image, and (2) multi-view (\textbf{MV}) by averaging the
center, 4 corners and their mirrored versions. In the first three
datasets, we evaluated our methods using two protocols, without
(\textbf{w/o. BBox}) and with (\textbf{w/. BBox}) the use of
ground-truth bounding box to crop out the object both at training and
testing.

\subsection{Stanford car dataset}

The first experiment validates our approach on the Stanford car
dataset~\cite{KrauseSDF013}, which contains $16,185$
images of $196$
car categories. We adopted the same $50$-$50$
split as in \cite{KrauseSDF013} by dividing the data into $8,144$
images for training and the rest for testing. Each category is
typically at the level of maker, model and year, \eg, Audi A5 Coupe
12. Following \cite{KrauseDSF013}, we manually assigned each
fine-grained label to one of $9$
coarse body types. Fig.~\ref{fig:exp_carsf}a summarizes the
distribution of training images for fine-grained and coarse labels.

Fig.~\ref{fig:exp_carsf}b-e compare between the original GN-SM and the
proposed GN-BGL$_m$ on a testing example. GN-SM confused the
ground-truth hatchback model (Acura Zdx Hatchback) with a very similar
sedan one (Acura TI Sedan). By jointly modeling with body type
(Fig.~\ref{fig:exp_carsf}c), however, GN-BGL$_m$ was able to predict
the correct fine-grained label. Fig.~\ref{fig:exp_carsf}f further
compares BGL with several previous works using different CNN
architectures. Without using CNN, the best published result, $69.5\%$,
was achieved in \cite{KrauseSDF013} by using a traditional LLC-based
representation. This number was beaten by ELLF \cite{KrauseGDLF14} by
learning more discriminative features using CNN. The bilinear model
\cite{LinCM15} recently obtained $88.2\%$ by combining two VGG
nets. By exploring the label dependency, the proposed BGL further
improves all the three CNN architectures using either single-view (SV)
or multi-view (MV) cropping. This consistent improvement demonstrates
BGLs advantage in modeling structure among fine-grained labels. Our
method of VGG-BGL$_m$ beats all previous works except
\cite{KrauseGDLF15}, which leveraged the part information in an
unsupervised way. However, we believe BGL can be combined with
\cite{KrauseGDLF15} to achieve better performance. In addition, BGL
has the advantage of predicting coarse label. For instance, GN-BGL$_m$
achieved $95.7\%$ in predicting the type of a car image.

\subsection{CUB-200-2011 dataset}

\begin{figure*}
  \centering
  \includegraphics[width=\textwidth]{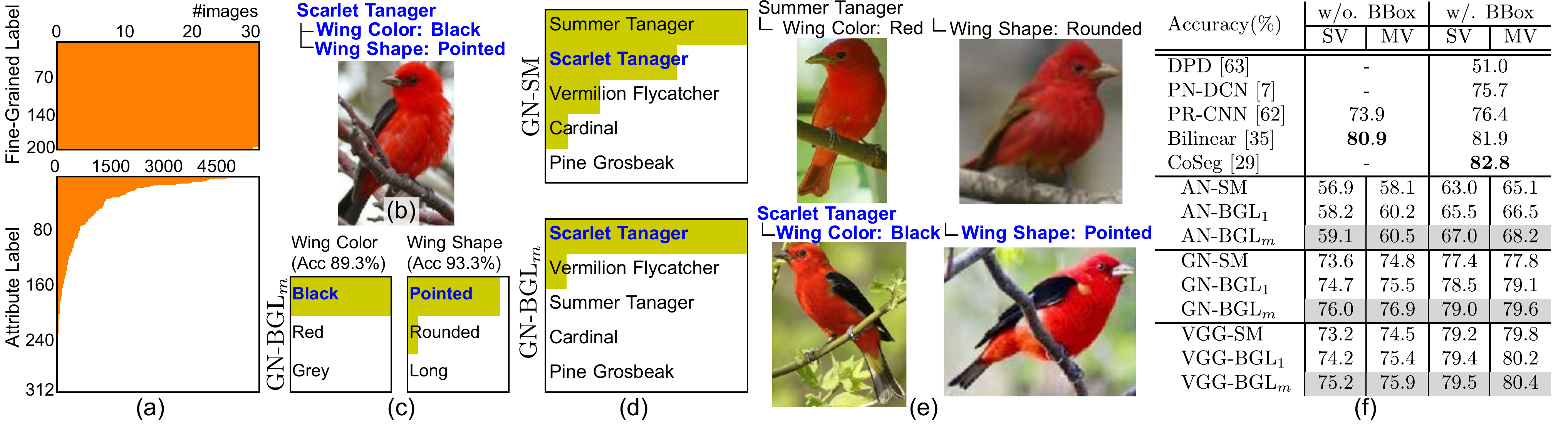}
  \caption{Comparison on CUB-200-2011 dataset. (a) The number of
    training images for each fine-grained category (top) and attribute
    (bottom). (b) An example of testing images. (c) 2 attributes
    predicted by GN-BGL$_m$. (d) Top-5 predictions generated by GN-SM
    approach (top) and the proposed BGL approach (bottom)
    respectively. (e) Similar training exemplars. (f) Accuracy.}
  \label{fig:exp_cub}
\end{figure*}

In the second experiment, we test our method on
CUB-200-2011~\cite{BransonHWPB14}, which is generally considered the
most competitive dataset within fine-grained recognition. CUB-200-2011
contains $11,788$ images of $200$ bird species. We used the provided
train/test split and reported results in terms of classification
accuracy. To get the label hierarchy, we adopted the annotated $312$
visual attributes associated with the dataset. See
Fig.~\ref{fig:exp_cub}a for the distribution of the fine-grained class
and attributes. These attributes are divided into $28$ groups, each of
which has about $10$ choices. According to the provided confidence
score, we assigned each attribute group with the most confident choice
for each bird specie. For instance, Fig.~\ref{fig:exp_cub}b shows an
example of Scarlet Tanger, most of which own black-color and
pointed-shape wings.

Fig.~\ref{fig:exp_cub}d compares the outputs between the baseline
GN-SM and the proposed GN-BGL$_m$. GN-SM predicted the bird in
Fig.~\ref{fig:exp_cub}b as a Summer Tanager, which is very close to
the actual class of Scarlet Tanager. By more closely comparing the
color and shape of the wings (Fig.~\ref{fig:exp_cub}c), however,
GN-BGL$_m$ decreased its possibility of being Summer Tanager. This
challenging example demonstrates the advantages of BGL in capturing
the subtle appearance difference between similar bird
species. Fig.~\ref{fig:exp_cub}f further compares BGL with
state-of-the-arts approaches in different settings. Early works such
as DPD~\cite{ZhangFID13} performed poorly when only using hand-crafted
features. The integration with CNN-based pose alignment techniques in
PN-DCN~\cite{BransonHPB14} and PR-CNN~\cite{ZhangDGD14} greatly
improve the overall performance. From the experiments, we observed
using BGL modules can consistently improved AN-SM, GN-SM and
VGG-SM. Without any pose alignment steps, our method GN-BGL$_m$
obtained $76.9\%$ without the use of bounding box, improving the
recent part-based method~\cite{ZhangDGD14} by $3\%$. In addition,
GN-BGL$_m$ achieved $89.3\%$ and $93.3\%$ accuracy on predicting
attributes of wing color and shape. However, our method still
performed worse than the latest methods \cite{LinCM15} and
\cite{KrauseGDLF15}, which show the significant advantage by exploring
part information for bird recognition. It is worth to emphasize that
BGL improves the last fully connected and loss layer for attribute
learning, while \cite{LinCM15} and \cite{KrauseGDLF15} focus on
integrating object part information into convolutional
layers. Therefore, it is possible to combine these two orthogonal
efforts to further improve the overall performance.



\subsection{Car-333 dataset}

\begin{figure*}
\centering
\includegraphics[width=\textwidth]{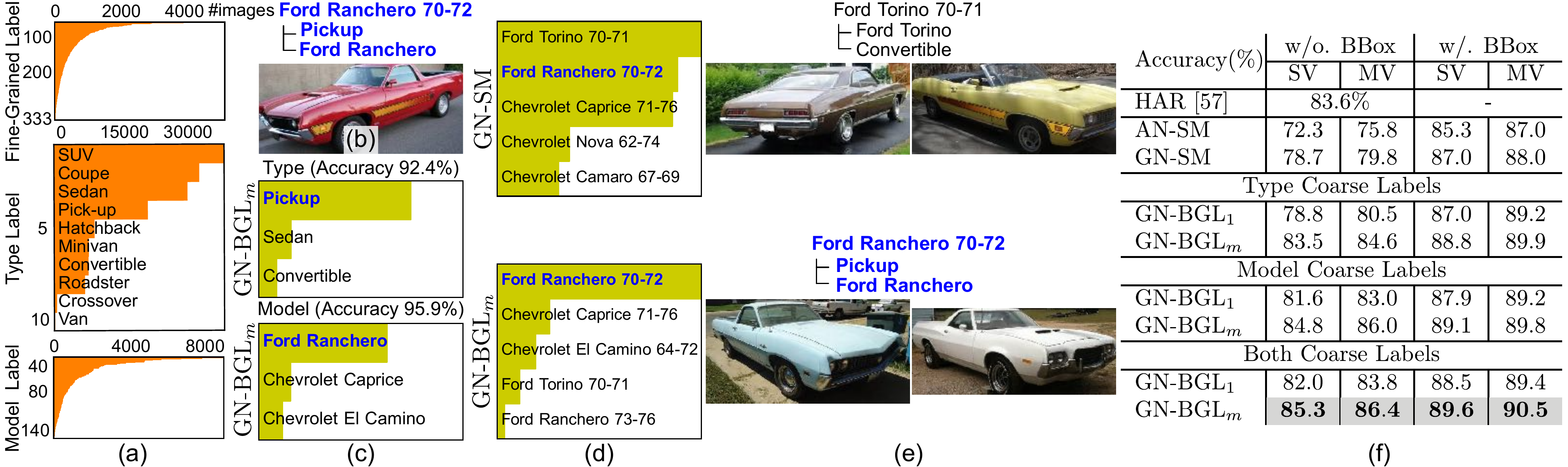}
\caption{Comparison on the Car-333 dataset. (a) Distribution of
  training images for each fine-grained (top) and coarse (bottom)
  category. (b) An example of testing image. (c) Type and model
  predicted by GN-BGL$_m$. (d) Top-5 prediction generated by GN (top)
  and the proposed BGL (bottom) respectively. (e) Similar training
  exemplars according to the input feature $\x$. (f) Accuracy.}
\label{fig:exp_car333}
\end{figure*}

In the third experiment, we test our method on the recently introduced
Car-333 dataset~\cite{XieYWL15}, which contains $157,023$ training
images and $7,840$ testing images. Compared to the Stanford car
dataset, the images in Car-333 were end-user photos and thus more
naturally photographed. Each of the $333$ labels is composed by maker,
model and year range. Notice that two cars of the same model but
manufactured in different year ranges are considered different
classes. To test BGL, we generated two sets of coarse labels: $10$
``type'' coarse labels manually defined according to the geometric
shape of each car model and $140$ ``model'' coarse labels by
aggregating year range labels. Please refer to
Fig.~\ref{fig:exp_car333} for the distribution of the training images
at each label level. The bounding box of each image was generated by
Regionlets~\cite{WangYZL13}, the state-of-the-art object detection
method.

Fig.~\ref{fig:exp_car333}b shows a testing example of Ford Ranchero
70-72. GN-SM recognized it as Ford Torino 70-71 because of the similar
training exemplars as shown in the top of
Fig.~\ref{fig:exp_car333}e. However, these two confused classes can be
well separated by jointly modeling the type and model probability in
GN-BGL$_m$. Fig.~\ref{fig:exp_car333}f summarizes the performance of
our method using different CNN architectures. The best published
result on this dataset was $83.6\%$ achieved by HAR~\cite{XieYWL15},
where the authors augmented the original training data with an
additional car dataset labeled view point information. We test BGL
with three combinations of the coarse labels: using either model or
type, and using model and type jointly. In particular, BGL gains much
more improvements using the $140$ model coarse labels than the $10$
type labels. This is because the images of the cars of the same
``model'' are more similar than the ones in the same ``type'' and it
defines richer relationships among fine-grained classes. Nevertheless,
BGL can still get benefit from putting the ``type'' labels on top of
the ``model'' labels to form a three-level label hierarchy. Finally,
GN-BGL$_m$ significantly improved the performance of GN-SM from
$79.8\%$ to $86.4\%$ without the use of bounding box. For more result
on AN and VGG, please refer to the supplementary material.

\begin{figure}
\centering
\includegraphics[width=.5\textwidth]{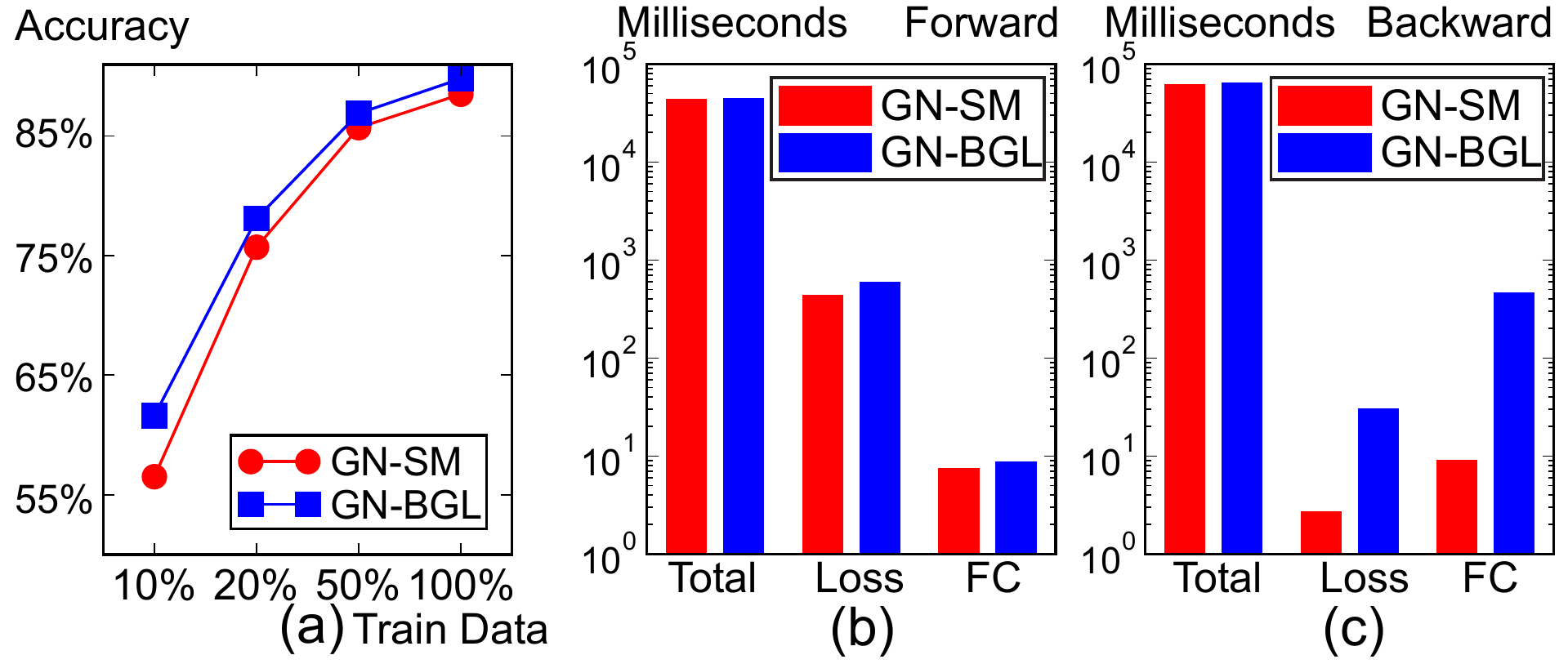}
\caption{More comparison on the Car-333 datasets. (a)
  Accuracy as a function of the amount of data used in training. (b)
  Time cost for forward passing given a $128$-image mini-batch, where
  Loss and FC denote the computation of the loss function and the last
  fully-connected layers respectively. (c) Time cost for backward
  passing.}
\label{fig:exp_car333_more}
\end{figure}

Since the Car-333 dataset is now big enough, we provide more
comparisons on the size of training data and time cost between GN-SM
and GN-BGL. Fig.~\ref{fig:exp_car333_more}a evaluates the performance
with respect to the different amounts of training data. BGL is able to
provide good improvement especially when training data is relatively
small. This is because the BGL formulation provides a way to
regularize CNN training to alleviate its overfitting
issue. Fig.~\ref{fig:exp_car333_more}b-c show the time cost for
performing forward and backward passing respectively given a
$128$-image mini-batch. Compared to GN-SM, GN-BGL needs only very
little additional computation to perform exact inference in the loss
function layer. This demonstrates the efficiency of modeling label
dependency in a bipartite graphs. For the last fully connected (FC)
layer, BGL performs exactly the same computation as GN in the forward
passing, but needs additional cost for updating the gradient
(Eq.~\ref{eq:opt_w1} and Eq.~\ref{eq:opt_w2}) in the backward
passing. Because both the loss and last FC layers take a very small
portion of the whole pipeline, we found the total time difference
between BGL and GN was minor.

\subsection{Food-975 dataset}

\begin{figure*}
\centering
\includegraphics[width=\textwidth]{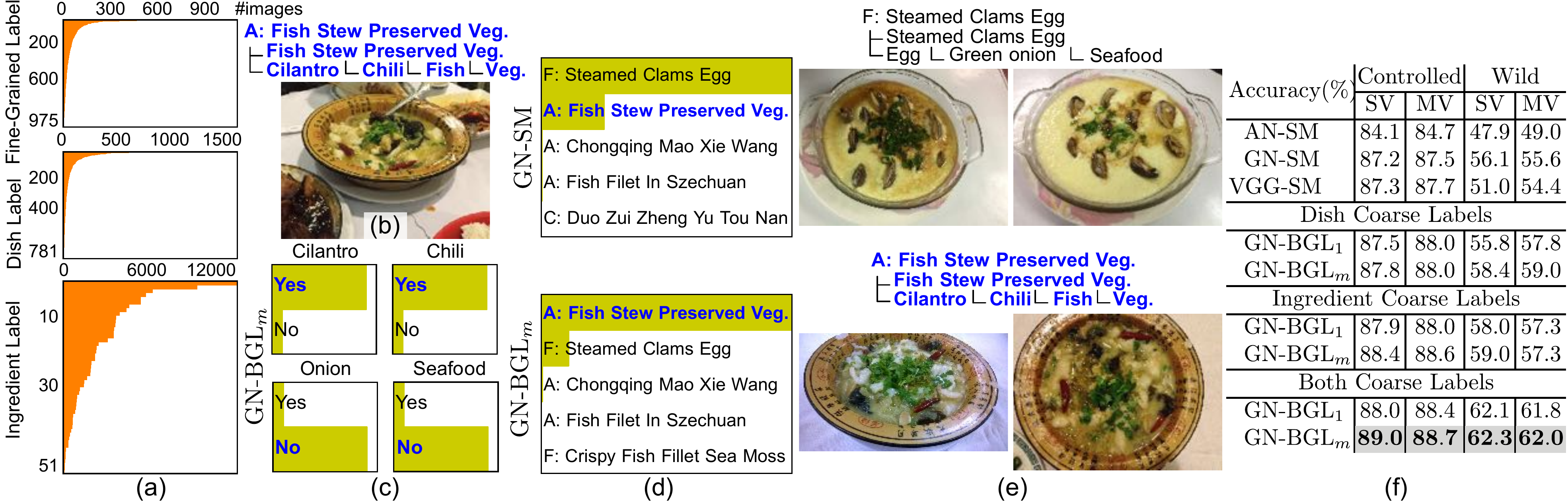}
\caption{Comparison on the Food-975 dataset. (a) Distribution of
  training images for each fine-grained (top), dish (middle), and
  ingredient (bottom) category. (b) An example of testing image. (c) 4
  ingredients predicted by GN-BGL$_m$.  (d) Top-5 predictions
  generated by GN-SM (top) and the proposed GN-BGL$_m$ (bottom)
  respectively. (e) Similar training exemplars according to the input
  feature $\x$. (f) Accuracy.}
\label{fig:exp_food975}
\end{figure*}

Now, let's come back to the task that we raised in the beginning of
the paper: given a food image from a restaurant, are we able to
recognize it as ``which restaurant which dish''? Apparently, this is
an ultra-fine grained image recognition problem and is very
challenging. As we mentioned before, such ultra-fine granularity
brings along rich relationships among the classes. In particular,
different restaurants may cook similar dishes with different
ingredients.


We collected a high quality food dataset. We sent 6 data collectors to
6 restaurants. Each data collector was in charge of one restaurant and
took the photo of almost every dish the restaurant had cooked during a
period of $1\sim2$ months. Finally, we captured $32,135$
high-resolution food photos of 975 menu items from the 6 restaurants
for training. We evaluated our method in two settings. To test in a
\emph{controlled} setting, we took additional $4951$ photos in
different days. To mimic a realistic scenario in the \emph{wild}, we
downloaded $351$ images from \url{yelp.com} posted by consumers
visiting the same restaurants. To model the class relationship, we
created a three-level hierarchy. In the first level, we have the $975$
fine-grained labels; in the middle, we created $781$ different dishes
by aggregating restaurant tags; at last, we came up a detailed list of
$51$ ingredient attributes\footnote{Find ingredient and restaurant
  list in the supplementary materials.} that precisely describes the
food composition.

Fig.~\ref{fig:exp_food975} compared the proposed BGL approach with
different baselines. Fig.~\ref{fig:exp_food975}e compares our method
with AN-SM, GN-SM and VGG-SM in both the controlled and wild
settings. We noticed that BGL approach consistently out-performed AN
and GN in both settings. This indicates the effectiveness of exploring
the label dependency in ultra-fine grained food
recognition. Interestingly, by using both the dish and ingredient
labels, BGL can gain a much larger improvement than only using one of
them. This implies the connections between the dish labels and the
ingredient labels have very useful information. Overall, by exploring
the label dependency, the proposed BGL approach achieved
$6 \sim 7\%$ improvement from GN baseline at the wild condition.


\section{Conclusion}

This paper proposed BGL to exploit the rich class relationships in the
very challenging ultra-fine grained tasks. BGL improves the
traditional softmax loss by jointly modeling fine-grained and coarse
labels through bipartite-graph labels. The use of a special bipartite
structure enables BGL to be efficient in inference. We also contribute
Food-975, an ultra-fine grained food recognition benchmark dataset. We
show that the proposed BGL approach improved previous work on a
variety of datasets.

There are several future directions to our work. (1) For the
ultra-fine grained image recognition, we may soon need to handle many
more classes. For example, we are constructing a large food dataset
from thousands of restaurants where the number of ultra-fine grained
food classes can grow into hundreds of thousands. We believe that the
research in this direction, ultra-fine-grained image recognition
(recognizing images almost on instance-level), holds the key for using
images as a media to retrieve information, which is often called
\emph{search by image}. (2) Although currently the label structure is
manually defined, it can potentially be learned during training (\eg,
\cite{SrivastavaS13}). On the other hand, we are designing a web
interface to scale up the attribute and ingredient labeling. (3) This
paper mainly discusses discrete labels in BGL. It is also interesting
to study the application with continuous labels (\eg, regression or
ranking). (4) Instead of operating only at class level, we plan to
generalize BGL to deal with image-level labels. This can make the
performance of BGL more robust in the case when the attribute label is
ambiguous for fine-grained class.

{\small
\bibliographystyle{ieee}
\bibliography{ref}
}

\section{Gradients of BGL's objective}

In the main submission, we have formulated the BGL's objective as
optimizing the following structural logistic likelihood over the
weights:
\begin{aligns} \label{eq:obj_bgl}
\min_{\W, \{\W_j\}_j} \ & \sum_{(\x, y) \in \cX} \Big(-\log
p_{y} - \sum_{j=1}^m \log p_{\phi^j_{y}}^j \Big) - \log
p_w,
\end{aligns}
where each components is computed as:

\begin{aligns}
p_i &= \frac{1}{z} \overbrace{e^{f_i} \prod_{j=1}^m
  e^{f^j_{\phi^j_{i}}}}^{h_i} \doteq \frac{1}{z} h_{i}, \label{eq:pi} \\
p_{c_j}^j &= \frac{1}{z} \sum_{i=1}^k g^j_{i c_j} h_{i}, \label{eq:pjc} \\
z & = \sum_{i=1}^k h_i, \label{eq:z} \\
p_w &= \prod_{i=1}^k \prod_{j=1}^m \prod_{c_j=1}^{k_j} e^{-
  \frac{\lambda}{2} g^j_{i c_j}  \| \w_i - \w_{c_j}^j \|^2}. \label{eq:pw}
\end{aligns}

Below we derived the gradients of each component.

\subsection{$\partial \log p_i / \partial f_{i'}$}

Given the definition of $p_i$ (Eq.~\ref{eq:pi}), if $i' = i$, then,
\begin{aligns}
  \frac{\partial \log p_i }{\partial f_{i'}}
  &= \frac{1}{p_i} \frac{h_{i} z - h_{i} h_{i'} }{z^2} \nonumber \\
  &= \frac{1}{p_i} (p_{i} - p_i p_{i'}) \nonumber \\
&= 1 - p_{i'}, \nonumber
\end{aligns}
otherwise,
\begin{aligns}
  \frac{\partial \log p_i }{\partial f_{i'}} &=
  \frac{1}{p_i} \frac{-h_i h_{i'}}{z^2} = -p_{i'}. \nonumber
\end{aligns}
Putting the result together, we have
\begin{aligns}
  \frac{\partial \log p_i }{\partial f_{i'}} &= 1_{[i'=i]} - p_{i'}. \nonumber
\end{aligns}

\subsection{$\partial \log p_{i} / \partial f_{c_j}^j$}

Given the definition of $p_i$ (Eq.~\ref{eq:pi}), if $g_{i c_j}^j = 1$
or equivalently $\phi^j_i = c_j$, then
\begin{aligns}
  \frac{\partial \log p_i }{\partial f^j_{c_j}}
  &= \frac{1}{p_i} \frac{h_i z - h_i \sum_{i'} g^j_{i' c_j} h_{i'} }{z^2} \nonumber \\
  &= \frac{1}{p_i} (p_i - p_i p_{c_j}^j) \nonumber \\
&= 1 - p_{c_j}^j, \nonumber
\end{aligns}
otherwise,
\begin{aligns}
  \frac{\partial \log p_i }{\partial f^j_c}
  &= \frac{1}{p_i} \frac{- h_i \sum_{i'} g^j_{i'c_j} h_{i'} }{z^2} = - p_{c_j}^j. \nonumber
\end{aligns}

Putting the result together, we have
\begin{aligns}
  \frac{\partial \log p_i }{\partial f_c^j} &= 1_{[g_{i c_j}^j=1]} - p_c^j. \nonumber
\end{aligns}

\subsection{$\partial \log p^j_{c_j} / \partial f_{i}$}

Given the definition of $p^j_{c_j}$ (Eq.~\ref{eq:pjc}), if $
g^j_{i c_j} = 1$ or equivalently $\phi^j_i = c_j$, then
\begin{aligns}
  \frac{\partial \log p^j_{c_j}}{\partial f_i}
  &= \frac{1}{p^j_{c_j}} \frac{h_i z - \sum_{i'} g^j_{i'c_j} h_{i'} h_i }{z^2} \nonumber \\
  &= \frac{1}{p^j_{c_j}} (p_i - p^j_{c_j} p_i) \nonumber \\
&= \frac{p_i}{p^j_{c_j}} - p_i, \nonumber
\end{aligns}
otherwise,
\begin{aligns}
  \frac{\partial \log p^j_{c_j}}{\partial f_i}
  &= \frac{1}{p^j_{c_j}} \frac{- \sum_{i'} g^j_{i'c_j} h_{i'} h_i }{z^2} = -p_i. \nonumber
\end{aligns}

Putting the result together, we have
\begin{aligns}
  \frac{\partial \log p_{c_j}^j }{\partial f_i} &= \frac{p_i}{p^j_{c_j}} 1_{[g_{ic_j}^j=1]} - p_i. \nonumber
\end{aligns}

\subsection{$\partial \log p^j_{c_j} / \partial f_{c_j'}^j$}

Given the definition of $p^j_{c_j}$ (Eq.~\ref{eq:pjc}), if $c_j' = c_j$, then
\begin{aligns}
  \frac{\partial \log p^j_{c_j} }{\partial f_{c_j'}^j}
  &= \frac{1}{p^j_{c_j}} \frac{\sum_i g_{i c_j}^j h_i z - \sum_i g_{i c_j}^j h_i
    \cdot \sum_{i} g_{i c_j'}^j h_{i}}{z^2} \nonumber \\
  &= \frac{1}{p^j_{c_j}}(p^j_{c_j} - p^j_{c_j} p^j_{c_j'}) \nonumber \\
  &= 1 - p^j_{c_j'}, \nonumber
\end{aligns}
otherwise,
\begin{aligns}
  \frac{\partial \log p^j_{c_j}}{\partial f_{c_j'}^j}
  &= \frac{1}{p^j_{c_j}} \frac{-\sum_i g^j_{i c_j} h_i \cdot \sum_i
    g^j_{i c_j'}
    h_i}{z^2} = -p^j_{c_j'} \nonumber.
\end{aligns}
Putting the result together, we have
\begin{aligns}
  \frac{\partial \log p_{c_j}^j }{\partial f_{c_j'}^j} &= 1_{[c_j' = c_j]} - p^j_{c_j'}. \nonumber
\end{aligns}

\subsection{$\partial \log p^j_{c_j} / \partial f_{c_l}^{l}, l \neq j$}
According to the definition of $p_{c_j}^j$ (Eq.~\ref{eq:pjc}), then
\begin{aligns}
  \frac{\partial \log p^j_{c_j} }{\partial f_{c_l}^{l}}
  &= \frac{1}{p^j_{c_j}} \frac{\sum_i g^j_{i c_j} g^l_{i c_l} h_i z -
    \sum_i g_{i c_j}^j h_i
    \cdot \sum_{i'} g_{i' c_l}^l h_{i'}}{z^2} \nonumber \\
  &= \frac{1}{p^j_{c_j}}(\sum_i g^j_{i c_j} g^l_{i c_l} p_i - p^j_{c_j} p^{l}_{c_l}) \nonumber \\
  &= \sum_i g^j_{i c_j} g^l_{i c_l} \frac{p_i}{p^j_{c_j}} - p^{l}_{c_l}. \label{eq:pjl}
\end{aligns}

\subsection{$\partial \log p_w / \partial \w_i$ and $\partial
  \log p_w / \partial \w_{c_j}^j$}

According to the definition of $p_w$ (Eq.~\ref{eq:pw}), we have
\begin{aligns}
  \log p_w &= \sum_{i=1}^k \sum_{j=1}^m \sum_{c_j=1}^{k_j} -
  \frac{\lambda}{2} g^j_{i c_j} \| \w_i - \w_{c_j}^j \|^2. \nonumber
\end{aligns}
Then,
\begin{aligns}
\frac{\partial \log p_w}{ \partial \w_i} = -\lambda \sum_{j=1}^m
\sum_{c_j=1}^{k_j} g^j_{i c_j}
(\w_i - \w_{c_j}^{j}). \nonumber
\end{aligns}
Similarly,
\begin{aligns}
\frac{\partial \log p_w}{ \partial \w_{c_j}^j} = - \lambda
\sum_{i=1}^k g^j_{i c_j}
(\w_{c_j}^j - \w_i). \nonumber
\end{aligns}

\section{Fast computation of gradient}

\begin{figure}
\centering
\includegraphics[width=.3\textwidth]{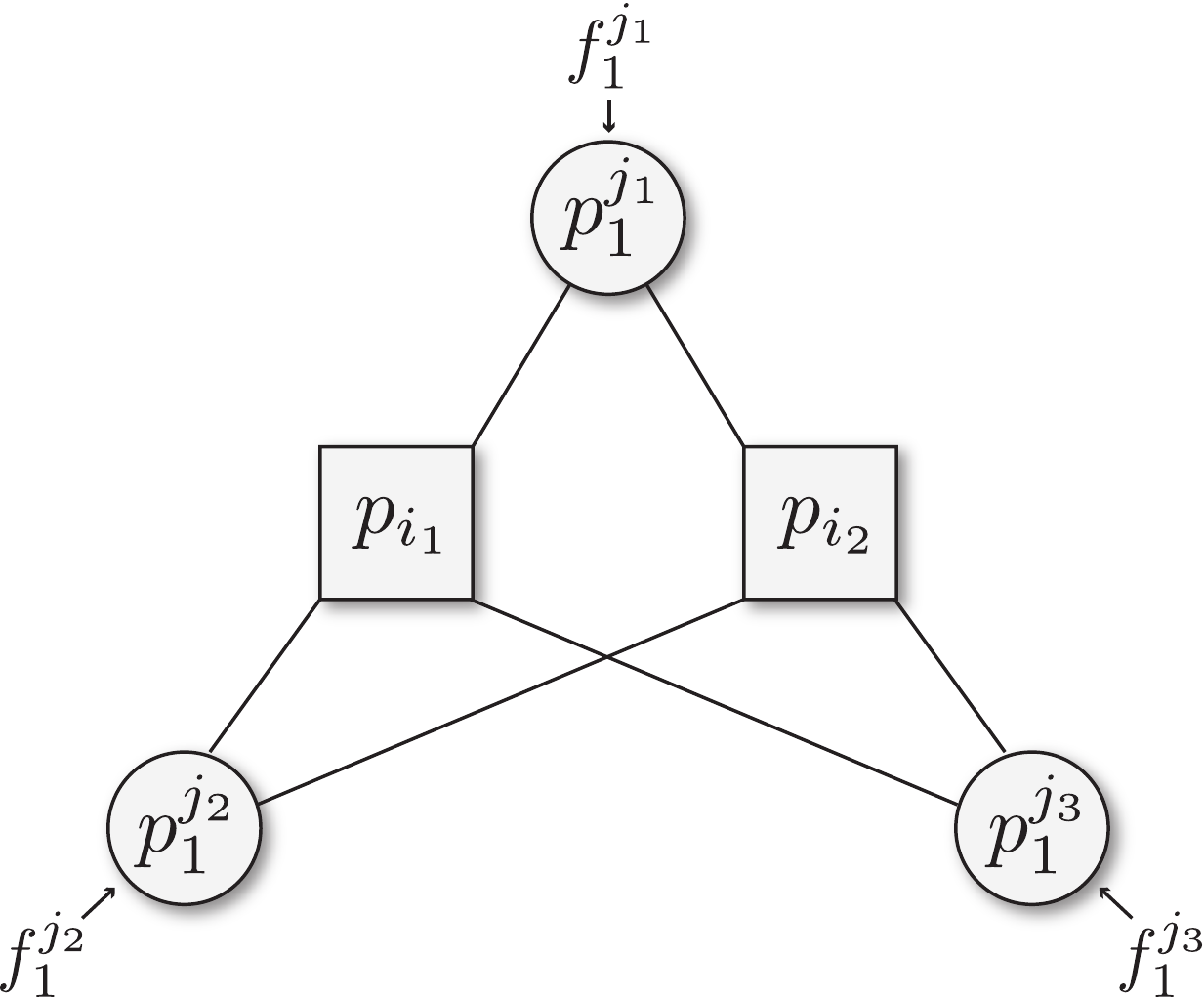}
\caption{A synthetic example for illustrating the fast computation of
  gradient, where the fine-grained labels $i_1$ and $i_2$ are
  connected with three coarse types $j_1$, $j_2$ and $j_3$ of size
  $1$.}
\label{fig:fast_gradient}
\end{figure}

Given a training sample of fine-grained class $y$ associated with $m$
coarse labels $\{\phi^j_y\}_{j=1}^m$, we need compute the
gradient $\partial \log p^j_{\phi^j_y} / \partial f_{c_l}^{l}$ for each
coarse label $\phi^j_y$ of type $j$ and the ones $c_l=1, \cdots, k_l$ of
any different type $l$, where $l \neq j$. As mentioned in the main
submission, computing the gradient directly using Eq.~\ref{eq:pjl} has
the large complexity of $O(km \sum_{j=1}^m k_j)$. Here we show how to
reduce the complexity by an order of magnitude to $O(km + k
\sum_{j=1}^m k_j)$.

To have a better understanding of the problem, let's consider the
synthetic example shown in Fig.~\ref{fig:fast_gradient}, where two
fine-grained labels $i_1$ and $i_2$ are connected with three coarse
types $j_1$, $j_2$ and $j_3$ of size $1$, \ie, $k_{j_1} = k_{j_2} =
k_{j_3} = 1$. Suppose that we have a training example of fine-grained
class $i_1$. According to Eq.~\ref{eq:pjl}, we can compute the
gradients for the input features $f^{j_1}_1$, $f^{j_2}_1$ and
$f^{j_3}_1$ respectively as,
\begin{aligns}
  \frac{\partial \log p_1^{j_2}}{\partial f^{j_1}_1} &=
  \frac{p_{i_1}}{p^{j_2}_1} + \frac{p_{i_2}}{p^{j_2}_1} - p^{j_1}_1, \quad
  \frac{\partial \log p_1^{j_3}}{\partial f^{j_1}_1} =
  \frac{p_{i_1}}{p^{j_3}_1} + \frac{p_{i_2}}{p^{j_3}_1} -
  p^{j_1}_1, \nonumber \\
  \frac{\partial \log p_1^{j_1}}{\partial f^{j_2}_1} &=
  \frac{p_{i_1}}{p^{j_1}_1} + \frac{p_{i_2}}{p^{j_1}_1} - p^{j_2}_1, \quad
  \frac{\partial \log p_1^{j_3}}{\partial f^{j_2}_1} =
  \frac{p_{i_1}}{p^{j_3}_1} + \frac{p_{i_2}}{p^{j_3}_1} - p^{j_2}_1,
  \nonumber \\
  \frac{\partial \log p_1^{j_1}}{\partial f^{j_3}_1} &=
  \frac{p_{i_1}}{p^{j_1}_1} + \frac{p_{i_2}}{p^{j_1}_1} - p^{j_3}_1, \quad
  \frac{\partial \log p_1^{j_2}}{\partial f^{j_3}_1} =
  \frac{p_{i_1}}{p^{j_2}_1} + \frac{p_{i_2}}{p^{j_2}_1} - p^{j_3}_1. \nonumber
\end{aligns}
In the final step of back propagation, we need to aggregate
the above gradients at each coarse label, that is,
\begin{aligns}
  \frac{\partial \log p_1^{j_2} + \partial \log p_1^{j_3}}{\partial
    f^{j_1}_1} &= \frac{p_{i_1}}{p^{j_2}_1} +
  \frac{p_{i_2}}{p^{j_2}_1} + \frac{p_{i_1}}{p^{j_3}_1} +
  \frac{p_{i_2}}{p^{j_3}_1} - 2 p^{j_1}_1, \label{eq:g1} \\
  \frac{\partial \log p_1^{j_1} + \partial \log p_1^{j_3}}{\partial
    f^{j_2}_1} &= \frac{p_{i_1}}{p^{j_1}_1} +
  \frac{p_{i_2}}{p^{j_1}_1} + \frac{p_{i_1}}{p^{j_3}_1} +
  \frac{p_{i_2}}{p^{j_3}_1} - 2 p^{j_2}_1, \label{eq:g2} \\
  \frac{\partial \log p_1^{j_1} + \partial \log p_1^{j_2}}{\partial
    f^{j_3}_1} &= \frac{p_{i_1}}{p^{j_1}_1} +
  \frac{p_{i_2}}{p^{j_1}_1} + \frac{p_{i_1}}{p^{j_2}_1} +
  \frac{p_{i_2}}{p^{j_2}_1} - 2 p^{j_3}_1. \label{eq:g3}
\end{aligns}
We found the above aggregation of gradients have many redundant
computations. For instance, by accumulating the scores
$\{p_i/p^j_1\}_{i,j}$ at the two fine-grained labels as,
\begin{aligns}
  q_{i_1} &= \frac{p_{i_1}}{p^{j_1}_1} + \frac{p_{i_1}}{p^{j_2}_1} +
  \frac{p_{i_1}}{p^{j_3}_1}, \quad
  q_{i_2} = \frac{p_{i_2}}{p^{j_1}_1} + \frac{p_{i_2}}{p^{j_2}_1} +
  \frac{p_{i_2}}{p^{j_3}_1}, \nonumber
\end{aligns}
we can compute the cumulative gradients in an alternative way:
\begin{aligns}
  \text{Eq.~\ref{eq:g1}} &= q_{i_1} - \frac{p_{i_1}}{p^{j_1}_1} +
  q_{i_2} - \frac{p_{i_2}}{p^{j_1}_1} - 2 p^{j_1}_1, \nonumber
  \\
  \text{Eq.~\ref{eq:g2}} &= q_{i_1} - \frac{p_{i_1}}{p^{j_2}_1} +
  q_{i_2} - \frac{p_{i_2}}{p^{j_2}_1} - 2 p^{j_2}_1, \nonumber
  \\
  \text{Eq.~\ref{eq:g3}} &= q_{i_1} - \frac{p_{i_1}}{p^{j_3}_1} +
  q_{i_2} - \frac{p_{i_2}}{p^{j_3}_1} - 2 p^{j_3}_1. \nonumber
\end{aligns}

Following this intuition, given a training sample of fine-grained
class $y$, we first pre-computed $k$ auxiliary variables that
accumulated all active\footnote{Each training example activates a
  sub-set ($m$) of all coarse labels ($\sum_j k_j$).} scores $p_i /
p^{j}_{\phi^j_y}$ at each fine-grained label, \ie,
\begin{aligns}
  q_i = \sum_{j=1}^m g_{i \phi^j_y}^j \frac{p_{i}}{p^{j}_{\phi^j_y}}, \nonumber
\end{aligns}
which takes $O(km)$ for computing all $\{q_i\}_{i=1}^k$. Then we
computed the cumulative gradient of each coarse label as,
\begin{aligns}
 \frac{\partial \sum_{j \neq l} \log p_{\phi^j_y}^j }{\partial
   f_{c_l}^l} = \sum_{i=1}^k \Big(q_i - \frac{p_i}{p^l_{c_l}} -
 p_{c_l}^{l} \Big). \label{eq:p_new}
\end{aligns}
Solving Eq.~\ref{eq:p_new} for all $\{f_{c_l}^l\}_{l,c_l}$ takes $O(k
\sum_{j=1}^{m}k_j)$. Therefore, the total complexity is $O(k m + k
\sum_{j=1}^{m}k_j)$.

\section{Car-333's VGG-BGL result}

\begin{figure}[b]
\centering
\includegraphics[width=.4\textwidth]{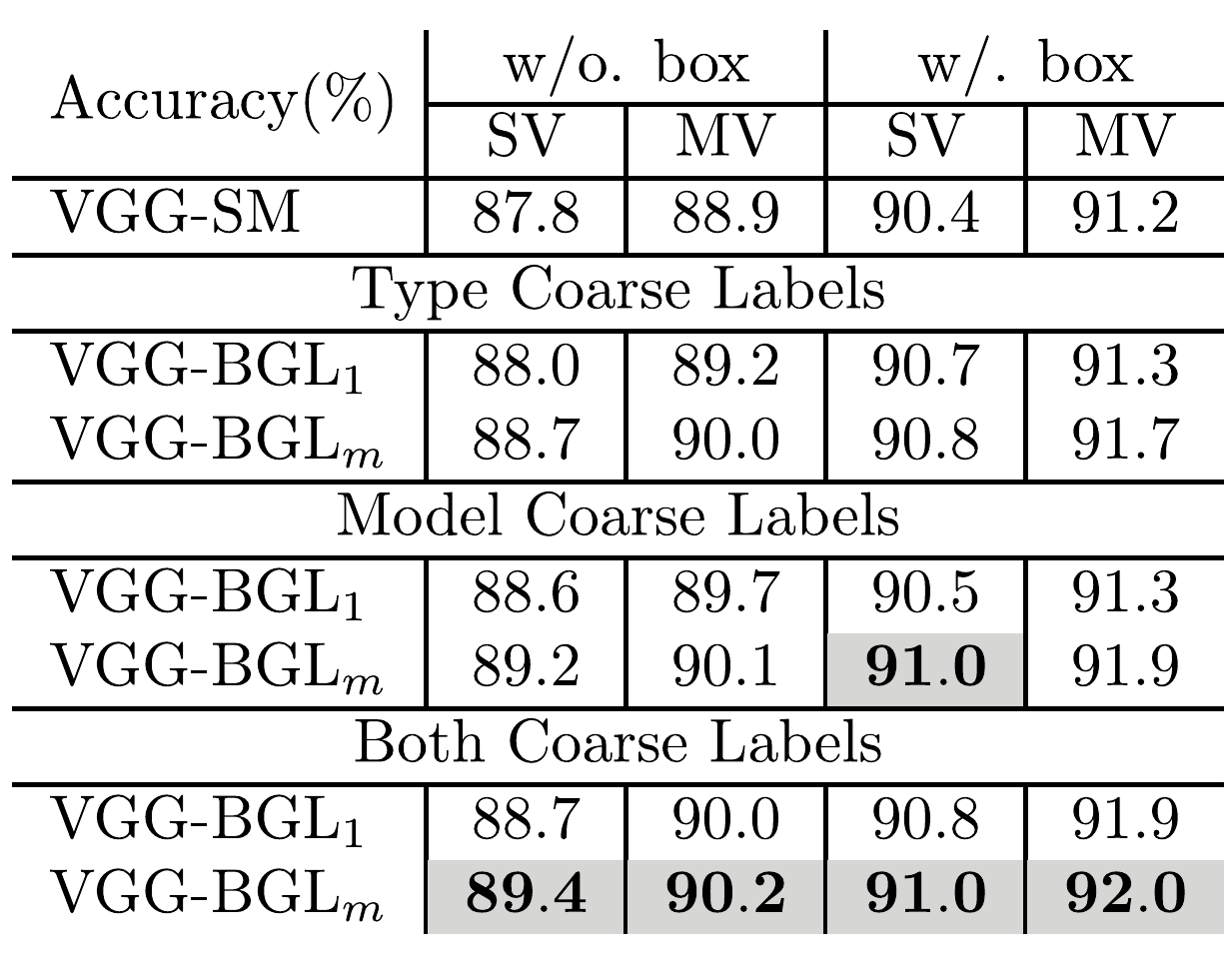}
\caption{Accuracy of Car-333 using VGG.}
\label{fig:exp_car333_vgg}
\end{figure}

Fig.~\ref{fig:exp_car333_vgg} summarizes the result of using BGL with
VGG for the Car-333 dataset.

\section{Food-975's restaurant list and ingredient labels}

Food-975 consists of $6$ Chinese restaurants in the bay area: Chef Yu,
Golden Garlic, Nutrition Restaurant, Lei Garden, Shanghai Dumpling and
Shanghai Restaurant.

\begin{figure*}
\centering
\includegraphics[width=1\textwidth]{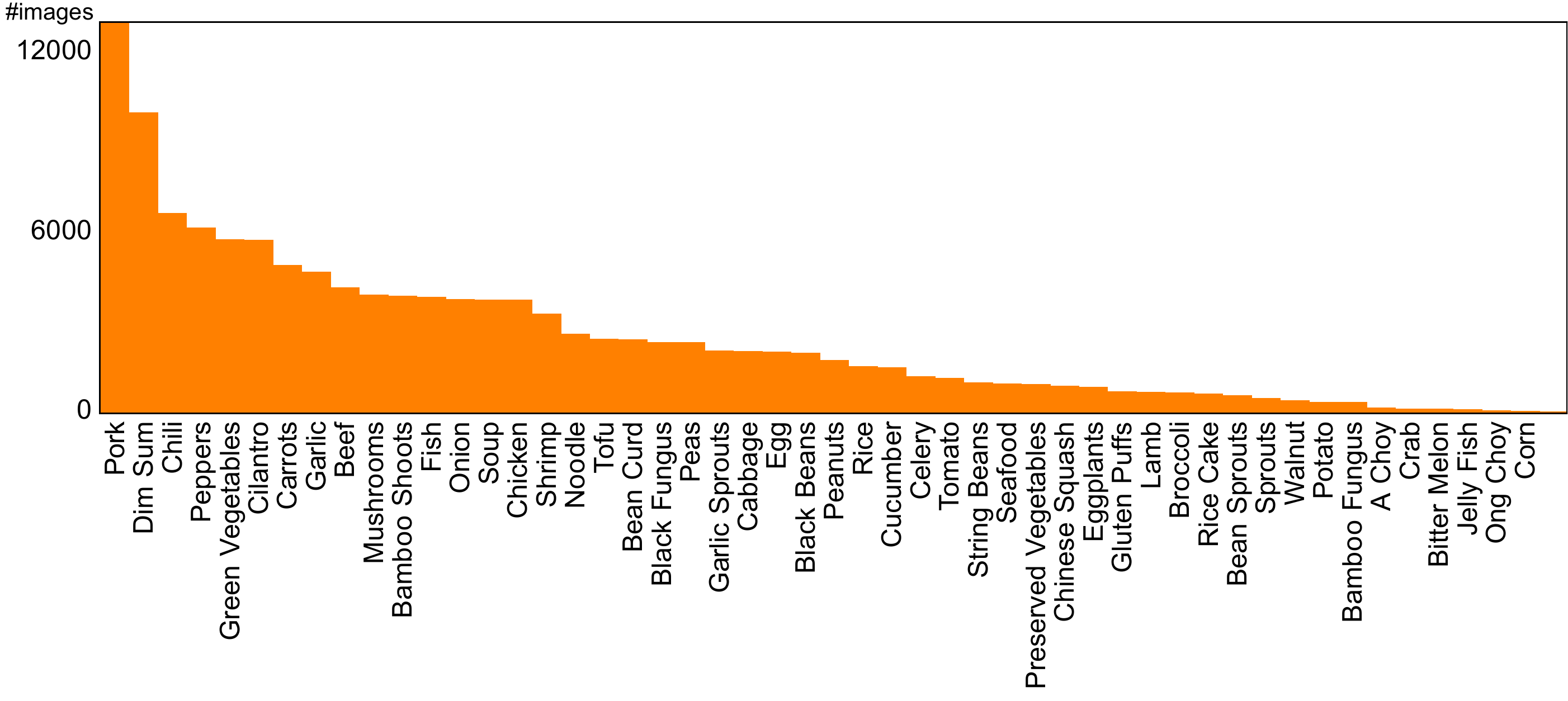}
\caption{Distribution of training images in Food-975 for all $51$
  ingredient attributes.}
\label{fig:exp_food975_in_ver}
\end{figure*}

Fig.~\ref{fig:exp_food975_in_ver} provides a detailed list of
ingredient labels for the new Food-975 dataset.

\section{More results}
\begin{figure*}
\centering
\includegraphics[width=\textwidth]{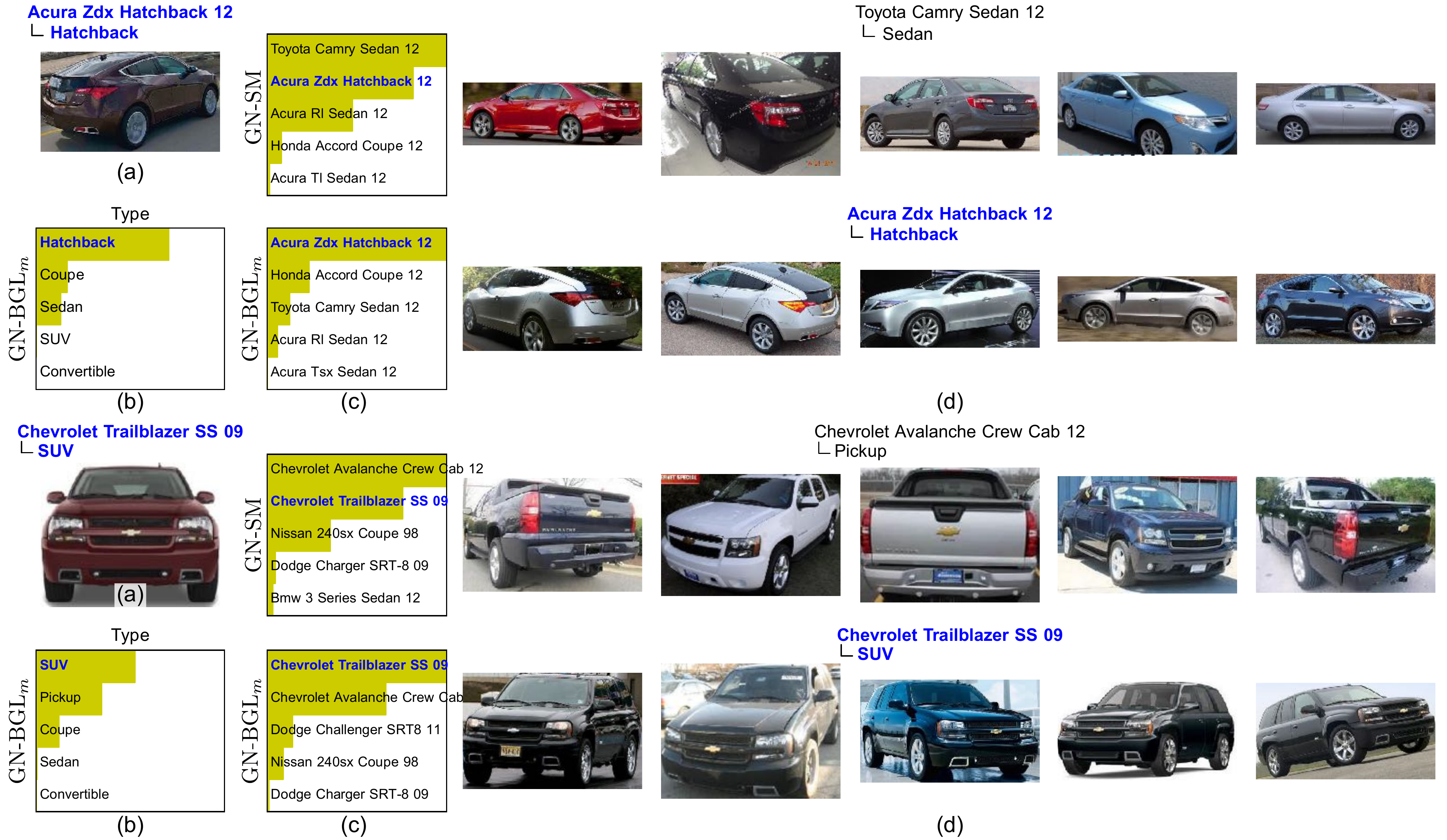}
\caption{More results on the Stanford car dataset. (a) An example of
  testing image. (b) Type and model predicted by GN-BGL$_m$. (c) Top-5
  predictions generated by GN-SM (top) and the proposed GN-BGL$_m$
  (bottom) respectively. (d) Similar training exemplars according to
  the input feature $\x$.}
\label{fig:exp_carsf_add}
\end{figure*}

\begin{figure*}
\centering
\includegraphics[width=\textwidth]{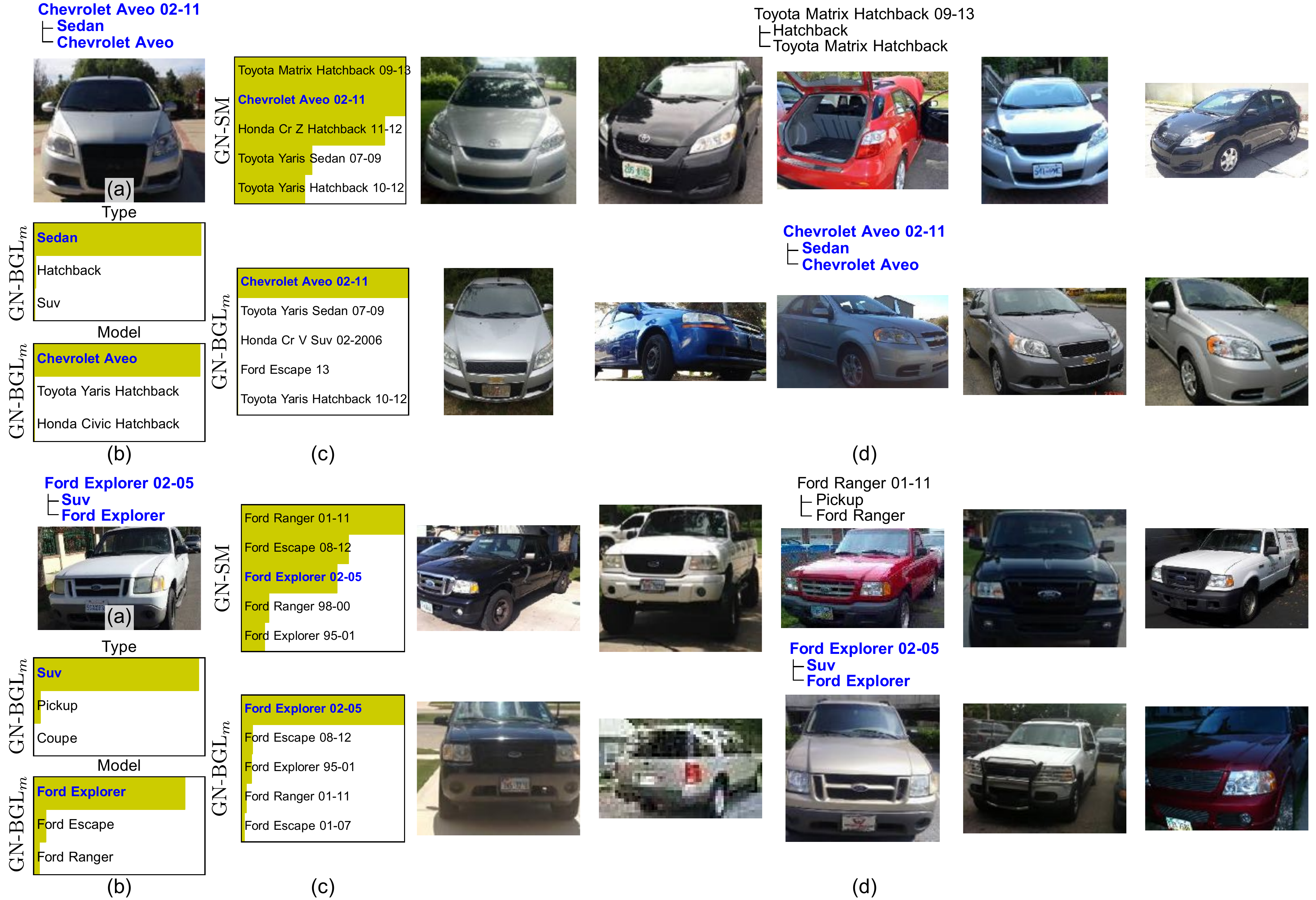}
\caption{More results on the Car-333 dataset. (a) An example of
  testing image. (b) Type and model predicted by GN-BGL$_m$. (c) Top-5
  predictions generated by GN-SM (top) and the proposed GN-BGL$_m$
  (bottom) respectively. (d) Similar training exemplars according to
  the input feature $\x$.}
\label{fig:exp_car333_add}
\end{figure*}

\begin{figure*}
\centering
\includegraphics[width=\textwidth]{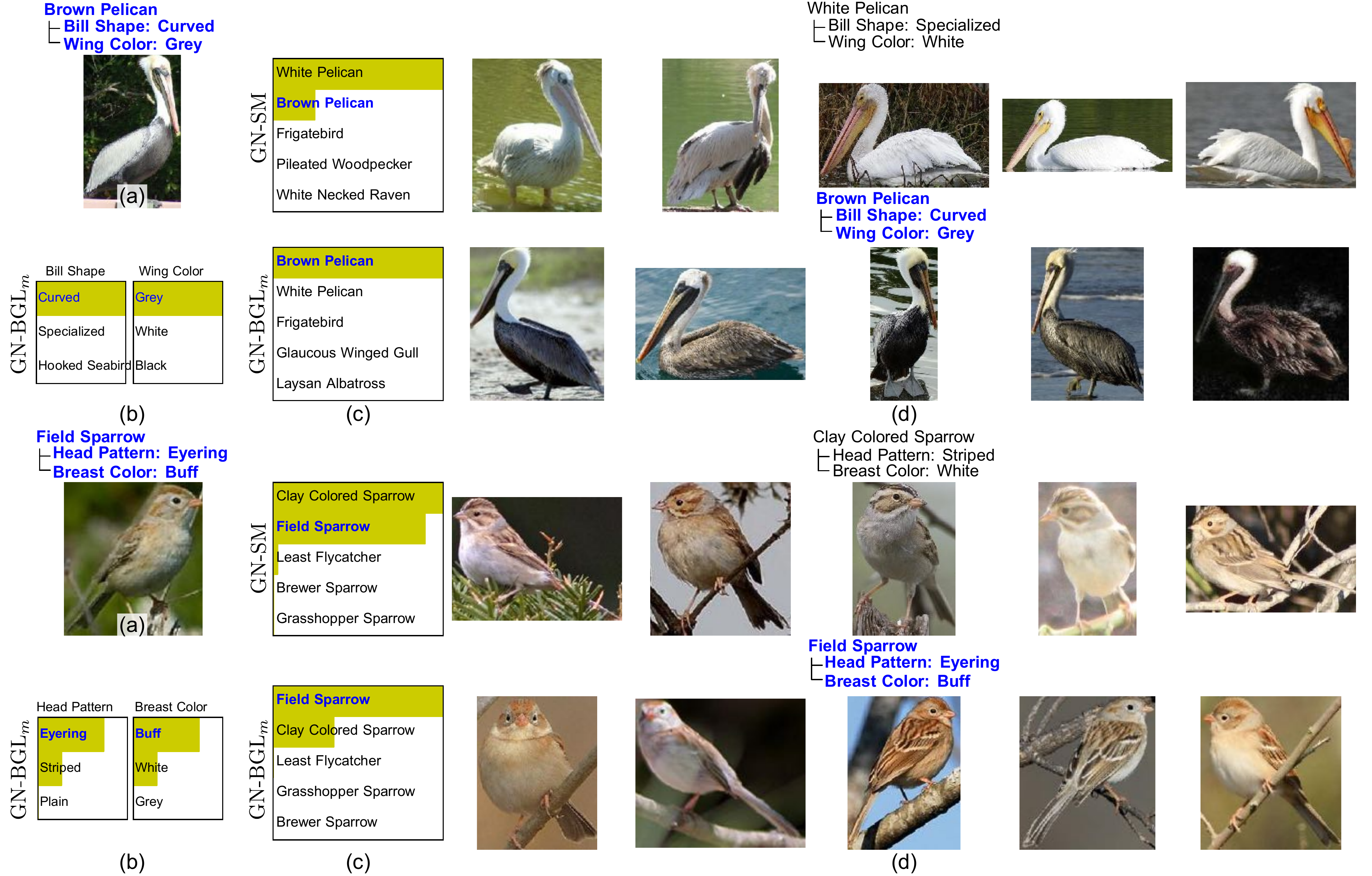}
\caption{More results on the CUB-200-2011 dataset. (a) An example of
  testing image. (b) Type and model predicted by GN-BGL$_m$. (c) Top-5
  predictions generated by GN-SM (top) and the proposed GN-BGL$_m$
  (bottom) respectively. (d) Similar training exemplars.}
\label{fig:exp_cub_add}
\end{figure*}

\begin{figure*}
\centering
\includegraphics[width=\textwidth]{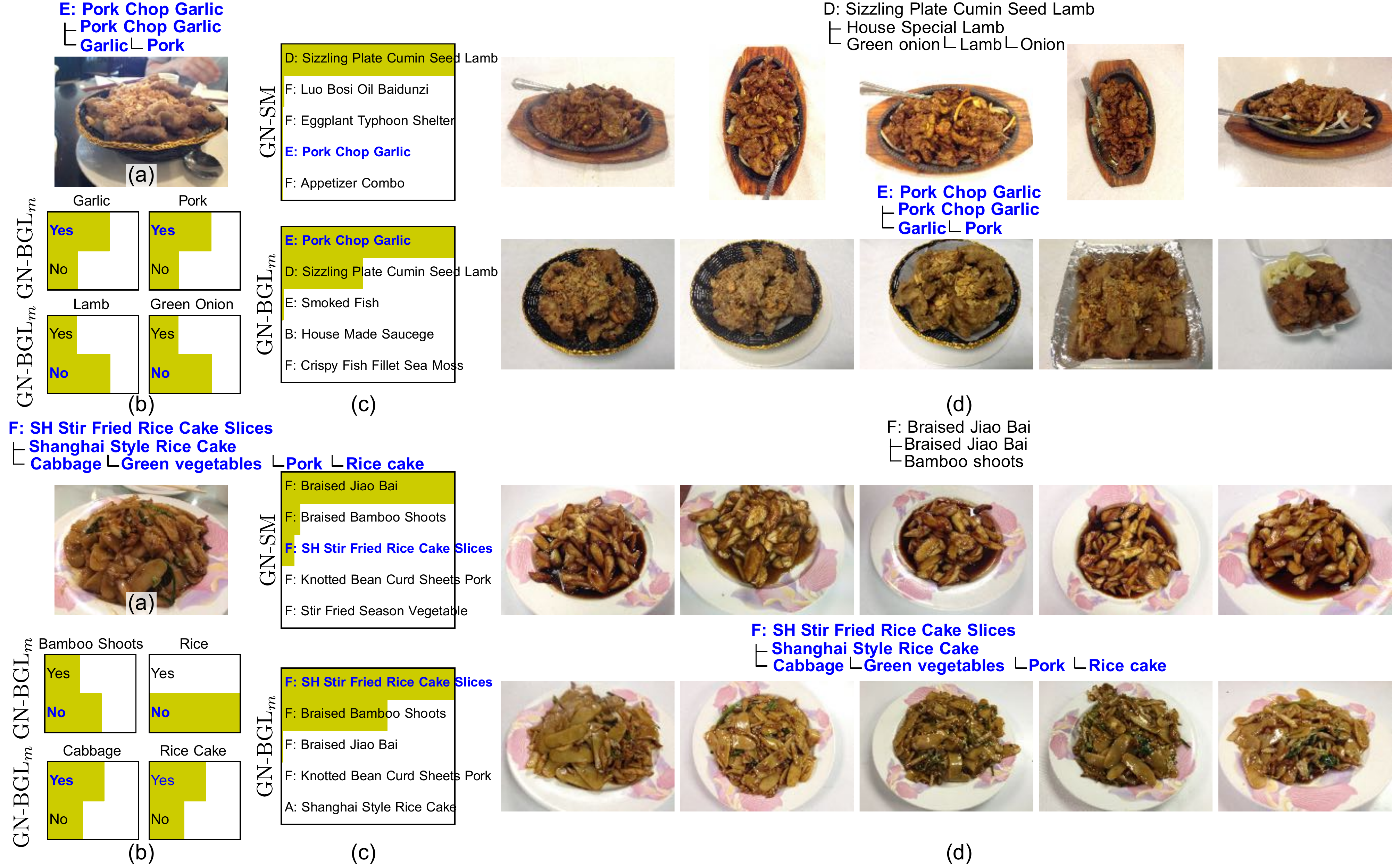}
\caption{More results on the Food-975 dataset. (a) An example of
  testing image. (b) Type and model predicted by GN-BGL$_m$. (c) Top-5
  predictions generated by GN-SM (top) and the proposed GN-BGL$_m$
  (bottom) respectively. (d) Similar training exemplars.}
\label{fig:exp_food975_add}
\end{figure*}

Fig.~\ref{fig:exp_carsf_add}-\ref{fig:exp_food975_add} provide
additional result for each dataset.

\end{document}